\documentclass[conference]{IEEEtran}
\IEEEoverridecommandlockouts
\usepackage{cite}
\usepackage{amsmath,amssymb,amsfonts}
\usepackage{algorithmic}
\usepackage{graphicx}
\usepackage{textcomp}
\usepackage{xcolor}
\def\BibTeX{{\rm B\kern-.05em{\sc i\kern-.025em b}\kern-.08em
    T\kern-.1667em\lower.7ex\hbox{E}\kern-.125emX}}
\usepackage{tabularx}
\usepackage{booktabs} 
\usepackage{pifont}
\usepackage{amsmath,amsfonts}
\usepackage{algorithmic}
\usepackage{algorithm}
\usepackage{array}
\usepackage{makecell}
\usepackage[caption=false,font=normalsize,labelfont=sf,textfont=sf]{subfig}
\usepackage{textcomp}
\usepackage{stfloats}
\usepackage{url}
\usepackage{verbatim}
\usepackage{graphicx}
\usepackage{cite}
\usepackage{bbm}
\usepackage{color}
\usepackage{cuted}
\usepackage{diagbox}
\usepackage[normalem]{ulem}
\usepackage{caption}
\usepackage{amssymb}
\usepackage{cite}
\usepackage{amsmath,amssymb,amsfonts}
\usepackage{algorithmic}
\usepackage[normalem]{ulem}
\usepackage{caption}
\usepackage{graphicx}
\usepackage{textcomp}
\usepackage{xcolor}
\usepackage{url}
\usepackage{bbm}
\usepackage{cuted}
\usepackage{algorithmic}
\usepackage{algorithm}

\begin{document}

\title{Optimized Certainty Equivalent Risk-Controlling Prediction Sets\\

\thanks{The work of J. Huang was supported by the King’s College London and China Scholarship Council for their Joint Full-Scholarship (K-CSC) (grant agreement No. 202206150005). The work of O. Simeone was supported by the European Union’s Horizon Europe Programme (grant agreement No. 101198347), by the Open Fellowship of the EPSRC (EP/W024101/1), and by the EPSRCproject (EP/X011852/1).}
}
\vspace{-2cm}
\author{\IEEEauthorblockN{Jiayi Huang\IEEEauthorrefmark{1}, Amirmohammad Farzaneh\IEEEauthorrefmark{1}, and Osvaldo Simeone\IEEEauthorrefmark{2}}
\IEEEauthorblockA{\IEEEauthorrefmark{1}Department of Engineering,  King’s College London, London, UK}
\IEEEauthorblockA{\IEEEauthorrefmark{2}Institute for Intelligent Networked
Systems (INSI), Northeastern University London, London, UK  \\
Email: \{jiayi.3.huang, amirmohammad.farzaneh\}@kcl.ac.uk, o.simeone@nulondon.ac.uk}\vspace{-1cm}}

\maketitle

\begin{abstract}
In safety-critical applications such as medical image segmentation, prediction systems must provide reliability guarantees that extend beyond conventional expected loss control. While risk-controlling prediction sets (RCPS) offer probabilistic guarantees on the expected risk, they fail to capture tail behavior and worst-case scenarios that are crucial in high-stakes settings. This paper introduces optimized certainty equivalent RCPS (OCE-RCPS), a novel framework that provides high-probability guarantees on general optimized certainty equivalent (OCE) risk measures, including conditional value-at-risk (CVaR) and entropic risk. OCE-RCPS leverages  upper confidence bounds to identify prediction set parameters that satisfy user-specified risk tolerance levels with provable reliability. We establish theoretical guarantees showing that OCE-RCPS satisfies the desired probabilistic constraint for loss functions such as miscoverage and false negative rate. Experiments on image segmentation demonstrate that OCE-RCPS consistently meets target satisfaction rates across various risk measures and reliability configurations, while OCE-CRC fails to provide probabilistic guarantees.
\end{abstract}



\begin{figure*}[t] \label{fig:overview}
    \noindent 
    \makebox[\textwidth]{\includegraphics[width=\textwidth]{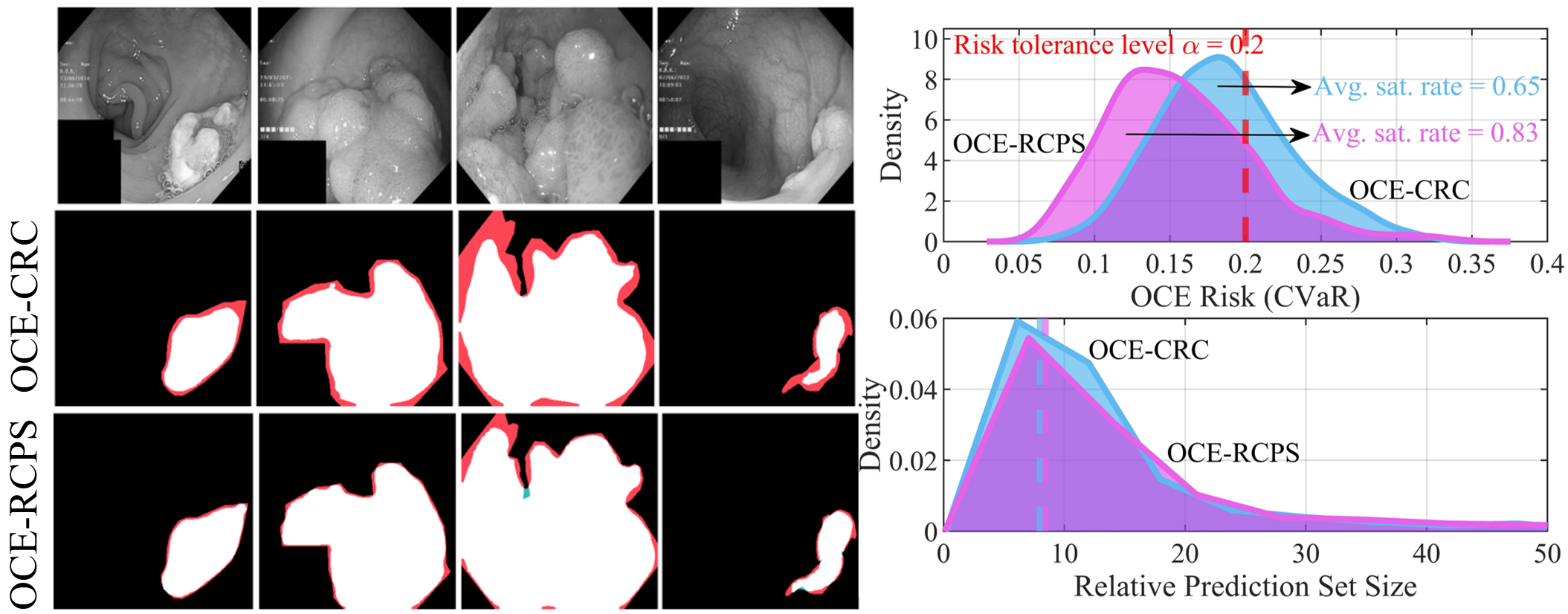}}
    \captionof{figure}{Given an input $x$, e.g., an image, and a pre-trained model $p(y|x)$, optimized certainty equivalent conformal risk control (OCE-CRC)  \cite{yeh2025conformal} and the proposed OCE risk-controlling prediction sets (OCE-RCPS) find a hyperparameter configuration $\hat{\lambda}$, e.g., an inclusion threshold, such that the reliability constraint $R_{\text{OCE}}(\hat{\lambda}) \leq \alpha$ is satisfied for any given OCE risk measure $R_{\text{OCE}}(\cdot)$, e.g., the conditional value-at-risk (CVaR). OCE-CRC \cite{yeh2025conformal} ensures this condition only on average with respect to the calibration data used to optimize the hyperparameter $\lambda$. In contrast, OCE-RCPS ensures that the condition $R_{\text{OCE}}(\hat{\lambda}) \leq \alpha$ holds with probability no smaller than a target level $1-\delta$. The left panel of the figure shows representative results for the CVaR of the false negative rate (FNR) loss in a segmentation task \cite{pogorelov2017kvasir}, with white, red, and green pixels denoting correct predictions, false negatives, and false positives, respectively. The right panel of the figure compares the empirical distribution of the CVaR and of the relative prediction set size with targets $\alpha=0.2$ and $1-\delta=0.8$. While OCE-CRC yields hyperparameters $\hat{\lambda}$ violating the requirement $R_{\text{OCE}}(\hat{\lambda}) \leq \alpha$ with probability larger than the requirement $\delta=0.2$, the proposed OCE-RCPS meets the target satisfaction rate $1-\delta=0.8$.}
\end{figure*}

\section{Introduction} \label{sec:intro}

\subsection{Motivation}

In safety-critical applications such as medical diagnosis, autonomous driving, and financial decision-making, predictive models must provide not only accurate predictions but also rigorous reliability guarantees. Traditional point predictions, even when accompanied by confidence scores from probabilistic models, often fail to quantify uncertainty reliably due to miscalibration \cite{guo2017calibration, huang2025calibrating}. This limitation can lead to catastrophic failures, such as missed cancer diagnoses in medical imaging \cite{abdar2021uncertainty} or false negatives in anomaly detection systems \cite{zhu2025conformal}. 

To address these challenges, recent advances in \emph{conformal prediction} (CP) have introduced prediction sets that provide distribution-free coverage guarantees \cite{shafer2008tutorial}. Among these methods, \emph{risk-controlling prediction sets} (RCPS) \cite{bates2021distribution} offer a principled approach to control the expected loss of the prediction set with high probability. Specifically, the loss, averaged over the test data point, is kept below a user-defined threshold with high probability with respect to the realization of the held-out dataset used for calibration. 
\vspace{-0.5cm}

However, RCPS focuses exclusively on the average risk, which fails to capture critical aspects of risk-sensitive decision-making, particularly the tail behavior of the loss distribution. As an example, consider the medical image segmentation task illustrated in Fig.~1. Given an input image  and a pre-trained segmentation model $p(y|x)$, the objective is to identify a set of pixels that constitute a reliable prediction of the object of interest. In this setting, controlling the {average} false negative rate (FNR), i.e., the average fraction of missed object pixels, may be insufficient, as medical practitioners often require stronger guarantees that account for worst-case test scenarios. A system that performs well on average,  but fails catastrophically on a non-negligible fraction of test data points may be unacceptable in clinical settings.

This motivates the need for optimized certainty equivalent (OCE) risk measures \cite{ben2007old}, which provide a flexible framework for controlling risk aversion through measures such as conditional value-at-risk (CVaR) and entropic risk. The recently introduced \emph{OCE conformal risk control} (OCE-CRC) \cite{yeh2025conformal} guarantees that the OCE risk remains below a target level $\alpha$. However, this guarantee holds only  on average over the possible held-out datasets used for calibration. As illustrated in Fig.~1, this implies that OCE-CRC can fail to produce prediction sets that satisfy the risk constraint for a high fraction of calibration datasets, leading to unreliable decisions in practice.

\begin{table}[t]
\centering
\caption{Comparison of prediction set calibration schemes: The table lists the types of guarantees provided with respect to (w.r.t.) test and calibration data.}\label{tab:1}
\label{tab:comparison}
\begin{tabular}{lcc}
\toprule
\textbf{Scheme} & \textbf{w.r.t. test data} & \textbf{w.r.t. calibration data} \\
\midrule
RCPS \cite{bates2021distribution} & Average & High probability \\
OCE-CRC \cite{yeh2025conformal} & OCE & Average \\
OCE-RCPS [this work] & OCE & High probability \\
\bottomrule
\end{tabular}
\end{table}




\subsection{Main Contributions}

In this work, we introduce OCE-RCPS, which extends RCPS to provide high-probability guarantees on general OCE risk measures (see Table \ref{tab:1}). By constructing upper confidence bounds (UCBs) on the OCE risk, OCE-RCPS ensures that the risk constraint  holds with high probability. This probabilistic guarantee enables practitioners to control the tail risk of their prediction systems with respect to both test data and calibration data, making OCE-RCPS particularly suitable for high-stakes applications. 


    

The remainder of this paper is organized as follows. Section~\ref{sec:problem_definition} formalizes the problem setting and introduces OCE risk measures. Section III reviews OCE-CRC as background. Section IV presents our proposed OCE-RCPS method along with theoretical guarantees. Section V provides experimental results on medical image segmentation. Finally, Section VI concludes the paper. 

\vspace{-0.2cm}

\section{Problem Definition} \label{sec:problem_definition}

Consider an inference setting, in which a pre-designed probabilistic model $p(y|x)$ is deployed to predict the target variable $y \in \mathcal{Y}$ given the input covariates $x \in \mathcal{X}$, where $p(y|x)$ is a conditional distribution in the target space $\mathcal{Y}$ given input $x$. In safety-critical applications, such as medical image segmentation, a conventional point prediction such as $\hat{y}(x) = \arg \max_{y \in \mathcal{Y}} p(y|x)$, can lead to missed diagnoses. Furthermore, the probabilistic distribution $p(y|x)$ is typically miscalibrated, and thus it cannot be directly used to quantify predictive uncertainty \cite{guo2017calibration, huang2025calibrating}. Calibration methods like CP address these limitations by leveraging held-out data with the goal of identifying a subset $\Gamma_{\lambda}(x)$ of the target domain $\mathcal{Y}$ that satisfies a user-specified reliability requirement.

The prediction set $\Gamma_{\lambda}(x)$ is constructed to include all output values $y \in \mathcal{Y}$ whose score $s(y|x)$ exceeds a threshold $1-\lambda$, i.e.,
\begin{align} \label{eq:set_construction}
    \Gamma_{\lambda}(x) = \{ y \in \mathcal{Y}: s(y|x) \geq 1 - \lambda \}.
\end{align}
The score $s(y|x)$ is a function of the predictive distribution $p(\cdot|x)$, with a typical example being the confidence level $s(y|x) = p(y|x)$ \cite{shafer2008tutorial}.

For example, as illustrated in Fig.~1, in medical image segmentation, the set $\mathcal{Y}$ contains all pixels in an image, and the prediction set $\Gamma_{\lambda}(x)$ corresponds to the collection of pixels whose score, representing the predicted probability of being positive exceeds, a threshold $1-\lambda$. Note that the prediction set (\ref{eq:set_construction}) satisfies the monotonicity condition
\begin{align} \label{eq:set_monotonicity}
    \Gamma_{\lambda_1}(x) \subseteq \Gamma_{\lambda_2}(x), \qquad \text{if $\lambda_1 \leq \lambda_2$},
\end{align}
with respect to the parameter $\lambda \in \mathbb{R}$.


The hyperparameter $\lambda$ controls the trade-off between the prediction size and reliability, and it is selected based on calibration data to meet target reliability requirements. In order to define the reliability requirements, let $\ell(y, \Gamma_{\lambda}(x))$ denote a non-negative bounded loss function that measures the discrepancy between the prediction set $\Gamma_{\lambda}(x)$ and output $y \in \mathcal{Y}$. We require that the loss function satisfies the monotonicity property
\begin{align} \label{eq:nesting_property}
    \ell(y, \Gamma_{\lambda_1}(x)) \geq \ell(y, \Gamma_{\lambda_2}(x)), \qquad \text{if $ \Gamma_{\lambda_1}(x) \subseteq \Gamma_{\lambda_2}(x)$ },
\end{align}
so that larger prediction sets correspond to lower losses. A canonical example is the miscoverage loss $\ell(y, \Gamma_{\lambda}(x)) = \mathbbm{1}(y \notin \Gamma_{\lambda}(x))$, where $\mathbbm{1}(\cdot)$ is the indicator function defined as $\mathbbm{1}(\text{true})=1$ and $\mathbbm{1}(\text{false})=0$. Other examples include the FNR in multi-label classification and image segmentation \cite{bates2021distribution}. 

In this work, we consider the general class of OCE risks, which include standard measures such as the expected loss, the entropic risk, and the CVaR (see Table~\ref{tab:oce_risks}). This family provides flexible and mathematically convenient metrics to control the level of risk aversion in the evaluation and optimization of decision policies \cite{ben2007old}. Formally, the class of OCE risks include metrics of the form
\begin{align} \label{eq:oce_risk}
    R_{\text{OCE}}(\lambda) = \inf_{t \in \mathbb{R}} \left\{ R(\lambda, t) =  t + \mathbb{E}[\phi \left(\ell(y, \Gamma_{\lambda}(x)) - t \right)] \right\},
\end{align}
where $\phi(\cdot)$ is a nondecreasing, closed, and convex cost function, and the inner expectation is taken with respect to the joint distribution $p(x,y)$ of the pair $(x,y)$. As summarized in Table~\ref{tab:oce_risks}, the choice of cost function $\phi(u)$ determines the specific risk measure induced by the OCE framework.

\begin{table}[t]
\centering
\caption{Examples of OCE risk measures.}\label{tab:2}
\label{tab:oce_risks}
\begin{tabular}{lc}
\toprule
\textbf{OCE risk measure} & \textbf{Cost function $\phi(u)$ in (\ref{eq:oce_risk})}\\
\midrule
Average risk & $\phi(u) = u$ \\
Entropic risk & $\phi(u) = \dfrac{1}{\beta} \left( e^{\beta u} - 1\right),  \beta > 0$ \\
CVaR  & $\phi(u) = \dfrac{1}{1 - \beta} \max(u, 0), \beta \in [0, 1)$ \\
\bottomrule
\end{tabular}
\end{table}

Given a calibration dataset $\mathcal{D}^{\text{cal}} = \{(x_i, y_i)\}_{i=1}^{|\mathcal{D}^{\text{cal}}|}$ with i.i.d. samples $(x,y) \sim p(x,y)$, we wish to identify a parameter $\hat{\lambda}$ such that the OCE risk (\ref{eq:oce_risk}) satisfies the reliability condition $R_{\text{OCE}}(\hat{\lambda}) \leq \alpha$ for some tolerance level $\alpha$ with probability no smaller than a user-defined level $1-\delta \in [0,1]$. This requirement is formally expressed as the inequality
\begin{align} \label{eq:oce_rcps_requirement}
    \Pr \big[ R_{\text{OCE}}(\hat{\lambda}) \leq \alpha  \big] \geq 1-\delta,
\end{align}
where the probability is taken over the distribution of the parameter $\hat{\lambda}$. When the OCE risk coincides with the average risk, the requirement (\ref{eq:oce_rcps_requirement}) coincides with the objective of RCPS \cite{yeh2025conformal}. Our goal is thus to generalize RCPS to any OCE risk.

\section{OCE Conformal Risk Control}
In this section, we briefly review OCE-CRC \cite{yeh2025conformal}, which provides guarantees on the average OCE risk. Specifically, given a calibration dataset $\mathcal{D}^{\text{cal}}$, OCE-CRC seeks a risk-controlled parameter $\hat{\lambda}$ such that the OCE risk $R_{\text{OCE}}(\hat{\lambda})$ does not exceed the target $\alpha$ on average with respect to the realizations of the calibration data $\mathcal{D}^{\text{cal}}$, i.e.,
\begin{align}\label{eq:oce-crc}
    \mathbb{E}\big[ R_{\text{OCE}}(\hat{\lambda}) \big] \leq \alpha.
\end{align}

To this end, for a fixed $t \in \mathbb{R}$, OCE-CRC selects the smallest value of the hyperparameter $\lambda$ such that an upper bound on the population OCE risk does not exceed a target risk tolerance level $\alpha$, i.e.,
\begin{align} \label{eq:optimal_lambda_crc}
    \hat{\lambda} = \inf \left\{ \lambda \in \mathbb{R}: \frac{|\mathcal{D}^{\text{cal}}|}{|\mathcal{D}^{\text{cal}}| + 1} \hat{R}^{\text{cal}}(\lambda,t) + \frac{B(\lambda, t)}{|\mathcal{D}^{\text{cal}}| + 1} \leq \alpha \right\},
\end{align}
where $\hat{R}^{\text{cal}}(\lambda,t)$ is the empirical estimate of $R(\lambda,t)$ in (\ref{eq:oce_risk}) evaluated on the calibration dataset $\mathcal{D}^{\text{cal}}$; $B(\lambda, t) = t + \phi (\ell_{\max}(\lambda) - t)$ is an upper bound on the function $R(\lambda,t)$ in (\ref{eq:oce_risk}), with $\ell_{\max}(\lambda) = \sup_{(x,y)} \ell(y, \Gamma_{\lambda}(x))$. As an example, for a bounded loss $\ell(y, \Gamma_{\lambda}(x)) \leq 1$, such as the FNR, we have $\ell_{\max}(\lambda) = 1$ for all values of $\lambda$.

While the guarantee (\ref{eq:oce-crc}) with the selected hyperparameter $\hat{\lambda}$ in (\ref{eq:optimal_lambda_crc}) holds for any fixed $t$, a poorly chosen parameter $t$ results in an unnecessarily larger prediction set $\Gamma_{\hat{\lambda}}(x)$. To improve the efficiency of the prediction set $\Gamma_{\hat{\lambda}}(x)$, OCE-CRC optimizes the parameter $t$ on a separate held-out optimization dataset $\mathcal{D}^{\text{opt}} = \{(x_i, y_i)\}_{i=1}^{|\mathcal{D}^{\text{opt}}|}$ with i.i.d. samples $(x,y) \sim p(x,y)$ by addressing the convex problem
\begin{align} \label{eq:optimal_t}
    t^* = \arg \min_{t \in \mathbb{R}} \hat{R}^{\text{opt}}(\lambda, t),
\end{align}
where $\hat{R}^{\text{opt}}(\lambda, t)$ is the empirical estimate of $R(\lambda,t)$ in (\ref{eq:oce_risk}) evaluated on the optimization dataset $\mathcal{D}^{\text{opt}}$. The use of a separate dataset $\mathcal{D}^{\text{opt}}$ ensures that the value $t^*$ in (\ref{eq:optimal_t}) is independent of the calibration data $\mathcal{D}^{\text{cal}}$, which is necessary to preserve the validity of the OCE risk guarantee in (\ref{eq:oce-crc}).

\section{OCE Risk-Controlling Prediction Set}

OCE-CRC guarantees the OCE risk requirement $R_{\text{OCE}}(\hat{\lambda}) \leq \alpha$ only on average with respect to calibration data as in (\ref{eq:oce-crc}), providing no control over the probability that the OCE risk constraint (\ref{eq:oce-crc}) is violated on any given trial. As discussed in Sec.~\ref{sec:intro}, in safety-critical applications such as medical image analysis, practitioners require assurance that the risk constraint is satisfied not merely on average, but with high probability with respect to the available calibration data (see Fig.~1). To describe the proposed solution, in this section, we first review RCPS \cite{bates2021distribution}, and then define OCE-RCPS.

\subsection{RCPS}
RCPS \cite{bates2021distribution} controls the expected loss $R_{\text{avg}}(\lambda) = \mathbb{E} \left[ \ell(y, \Gamma_{\lambda}(x)) \right]$, which corresponds to a special case of the OCE risk (see Table~\ref{tab:oce_risks}). Specifically, RCPS identifies a hyperparameter $\hat{\lambda}$ such that condition (\ref{eq:oce_rcps_requirement}) is satisfied for the expected loss $R_{\text{avg}}(\hat{\lambda})$. To this end, using the calibration dataset $\mathcal{D}^{\text{cal}}$, RCPS constructs an UCB  $\hat{R}^+_{\text{avg}}(\lambda, \delta)$ on the expected loss $R_{\text{avg}}(\lambda)$, satisfying the inequality $R_{\text{avg}}(\lambda) \leq \hat{R}^+_{\text{avg}}(\lambda, \delta)$ with probability at least $1-\delta$, i.e.,
\begin{align} \label{eq:rcps_inequality}
    \Pr\big[ R_{\text{avg}}(\lambda) \leq \hat{R}^+_{\text{avg}}(\lambda, \delta) \big] \geq 1 - \delta.
\end{align}

Then, RCPS selects the smallest value of hyperparameter $\lambda$ such that the UCB does not exceed the target value $\alpha$ for all $\lambda' \geq \lambda$, i.e.,
\begin{align} \label{eq:optimal_lambda_rcps}
    \hat{\lambda} = \inf \left\{ \lambda \in \mathbb{R}: \hat{R}_{\text{avg}}^+(\lambda', \delta) \leq \alpha \text{ for all } \lambda' \geq \lambda   \right\}. 
\end{align}

\subsection{OCE-RCPS} \label{sec:oce_rcps}
OCE-RCPS extends RCPS to offer control over any OCE risk. For any fixed value of parameter $t \in \mathbb{R}$, the definition of the OCE (\ref{eq:oce_risk}) implies the inequality 
\begin{align} \label{eq:inequality_t}
    R_{\text{OCE}}(\lambda) \leq R(\lambda, t).
\end{align}
The quantity $R(\lambda, t)$ in (\ref{eq:oce_risk}) can be estimated using the calibration dataset $\mathcal{D}^{\text{cal}}$. Specifically, in a manner similar to RCPS, we use the calibration dataset $\mathcal{D}^{\text{cal}}$ to construct an UCB on the function $R(\lambda, t)$. Here we adopt the state-of-the-art Waudby-Smith and Ramdas (WSR) UCB \cite{bates2021distribution}.

Given the calibration dataset $\mathcal{D}^{\text{cal}}$ and a target probability $\delta$, the WSR UCB on the risk $R(\lambda,t)$ is given by
\begin{align} \label{eq:wsr_bound}
    \hat{R}^{+}(\lambda, t, \delta) =\inf\left\{
    R\ge 0:\;
    \max_{i \in \mathcal{D}^{\text{cal}}}\mathcal{K}_i(R, \lambda)> \frac{1}{\delta},
    \right\}
\end{align}
where function $\mathcal{K}_i(R,\lambda)$ is defined as
\begin{align}
    \mathcal{K}_i(R,\lambda) =\prod_{j=1}^{i} \left( 1+\eta_j\Bigl(t + \phi\bigl(\ell(y_j,\Gamma_{\lambda}(x_j))-t\bigr) - R\Bigr) \right),
\end{align}
with $\eta_j > 0$. The constant $\eta_j$ can be potentially optimized as a function of the previous $j-1$ observations $(x_i,y_i)_{i=1}^{j-1}$ in the calibration dataset (see \cite[Prop.~5]{bates2021distribution}). For our example, we adopt the online Newton step scheme \cite{bates2021distribution}. The WSR UCB in (\ref{eq:wsr_bound}) satisfies the inequality
\begin{align} \label{eq:single_ucb}
    \Pr \big[ R(\lambda, t) \leq \hat{R}^+(\lambda, t, \delta) \big] \geq 1 - \delta,
\end{align}
where the probability is taken over the calibration data.

The following lemma establishes that the UCB $\hat{R}^+(\lambda, t, \delta)$ serves as a valid UCB on the OCE risk $R_{\text{OCE}}(\lambda)$ for any $t\in\mathbb{R}$.

\textbf{Lemma 1:} \textit{For any fixed value of parameter $t \in \mathbb{R}$, the UCB $\hat{R}^+(\lambda, t, \delta)$ satisfies an inequality
\begin{align} \label{eq:lemma}
    \Pr \big[ R_{\text{OCE}}(\lambda) \leq \hat{R}^+(\lambda, \delta, t) \big] \geq 1 - \delta.
\end{align}
Proof:} \text{Use (\ref{eq:inequality_t}) in (\ref{eq:single_ucb})}. \hfill $\blacksquare$

Finally, OCE-RCPS chooses the smallest value of hyperparameter $\lambda$ such that the UCB in (\ref{eq:wsr_bound}) does not exceed the target $\alpha$ for all $\lambda' \geq \lambda$, i.e.,
\begin{align} \label{eq:optimal_lambda_oce_rcps}
    \hat{\lambda} = \inf \left\{ \lambda \in \mathbb{R}: \hat{R}^+(\lambda', \delta, t) \leq \alpha \text{ for all } \lambda' \geq \lambda   \right\}.
\end{align}
As for OCE-CRC, the tightness of the resulting set $\Gamma_{\hat{\lambda}}(x)$ depends on the choice of hyperparameter $t$. As in OCE-CRC, we propose to choose an optimized value of parameter $t$ as in (\ref{eq:optimal_t}) based on held-out data $\mathcal{D}^{\text{opt}}$.

\subsection{Theoretical Guarantees}
OCE-RCPS satisfies the desired reliability requirement (\ref{eq:oce_rcps_requirement}).

\textbf{Theorem 1:} \textit{Let $\hat{\lambda}$ be selected as in (\ref{eq:optimal_lambda_oce_rcps}), with parameter $t^*$ in (\ref{eq:optimal_t}). Under the monotonicity condition (\ref{eq:set_monotonicity}), OCE-RCPS satisfies the requirement
\begin{align}
    \Pr \big[ R_{\text{OCE}}(\hat{\lambda}) \leq \alpha  \big] \geq 1-\delta.
\end{align}}


\section{Experiments}
In this section, to validate the proposed OCE-RCPS scheme, we report empirical results for the tumor segmentation task \cite{angelopoulos2022conformal}.

\subsection{Task, Baselines, and Implementation}
Following \cite{bates2021distribution}, we aggregate data from five open-source polyp segmentation benchmarks: Kvasir \cite{pogorelov2017kvasir}, Hyper-Kvasir \cite{borgli2020hyperkvasir}, CVC-ClinicDB and CVC-ColonDB \cite{bernal2012towards}, and ETIS-Larib \cite{silva2014toward}, yielding a combined dataset with $1781$ examples of segmented polyps. We adopt a pre-trained PraNet \cite{fan2020pranet} as the base segmentation model $p(y|x)$.

For all schemes and for each trial, we randomly select a reference dataset $\mathcal{D}$ of size $|\mathcal{D}| = 1000$ and a test dataset $\mathcal{D}^{\text{te}}$ of size $|\mathcal{D}^{\text{te}}| = 781$. The reference dataset $\mathcal{D}$ is randomly split into two disjoint datasets, namely the optimization dataset $\mathcal{D}^{\text{opt}}$ used to estimate the parameter $t^*$, and the calibration dataset $\mathcal{D}^{\text{cal}}$ used to identify the parameter $\hat{\lambda}$, with cardinalities $|\mathcal{D}^{\text{opt}}| = 200$ and $|\mathcal{D}^{\text{cal}}| = 800$, respectively. All results are averaged over $1000$ independent trials\footnote{Code can be found at \url{https://github.com/kclip/OCE-RCPS}.}.

We adopt the FNR loss, i.e., $\ell(y, \Gamma_{\lambda}(x)) = 1-|y \cap \Gamma_{\lambda}(x)|/|y|$, where $y$ is the set f pixels containing the object of interest and $|y|$ its cardinality, and consider the following evaluation metrics:
\begin{itemize}
    \item \emph{Average satisfaction rate}: This is the proportion of trials (i.e., calibration datasets) in which the OCE risk $R(\hat{\lambda})$ of the selected hyperparameter $\hat{\lambda}$ does not exceed the target tolerance level $\alpha$. This quantity estimates the left-hand side of the requirement (\ref{eq:oce_rcps_requirement}).
    \item \emph{Relative prediction set size}: This is the ratio, $|\Gamma(\hat{\lambda})|/|y|$, between the size of the prediction set and the ground-truth size of the polyp region.
    \item \emph{Distributions of the OCE risk metric and the relative prediction set size}: The distributions of the OCE risk metric and of the relative prediction set size are obtained as empirical estimates using kernel density estimation across the different trials.
\end{itemize}

\subsection{Results}

\noindent \textbf{OCE-CRC versus OCE-RCPS:}
To start, in the right top panel of Fig.~1 and top panel of Fig.~\ref{fig:dist_entropic}, we report the performance of OCE-CRC \cite{yeh2025conformal} and of the proposed OCE-RCPS in terms of risk and of relative prediction set size, for the CVaR with parameter $\beta=0.9$ in the right panel of Fig.~1 and for the entropic risk with parameter $\beta = 3$ in Fig.~\ref{fig:dist_entropic}. As shown in the right top panel of Fig.~1 and top panel of Fig.~\ref{fig:dist_entropic}, the distribution of the CVaR and of the entropic risk obtained by OCE-CRC exhibits significant probability mass beyond the target risk tolerance level $\alpha=0.2$, resulting in an average satisfaction rate of only $0.65$ and of only $0.58$, respectively, well below the target of $1-\delta = 0.8$. This confirms that controlling only the expected risk, as in OCE-CRC, provides no guarantee on the tail behavior of the risk distribution. In contrast, OCE-RCPS achieves an average satisfaction rate of $0.83$ and of $0.93$ for the CVaR and entropic risk, respectively, meeting the target reliability requirement as in (\ref{eq:oce_rcps_requirement}). To achieve these results, as seen in the bottom panel of the figures, OCE-RCPS increases the median of relative prediction set size from $7.98$ to $8.45$ for the CVaR, and from $1.74$ to $2.74$ for the entropic risk, respectively.


\setcounter{figure}{1}
\vspace{-0.4cm}
\begin{figure} [htb] 
    \centering
    \centerline{\includegraphics[width=\linewidth]{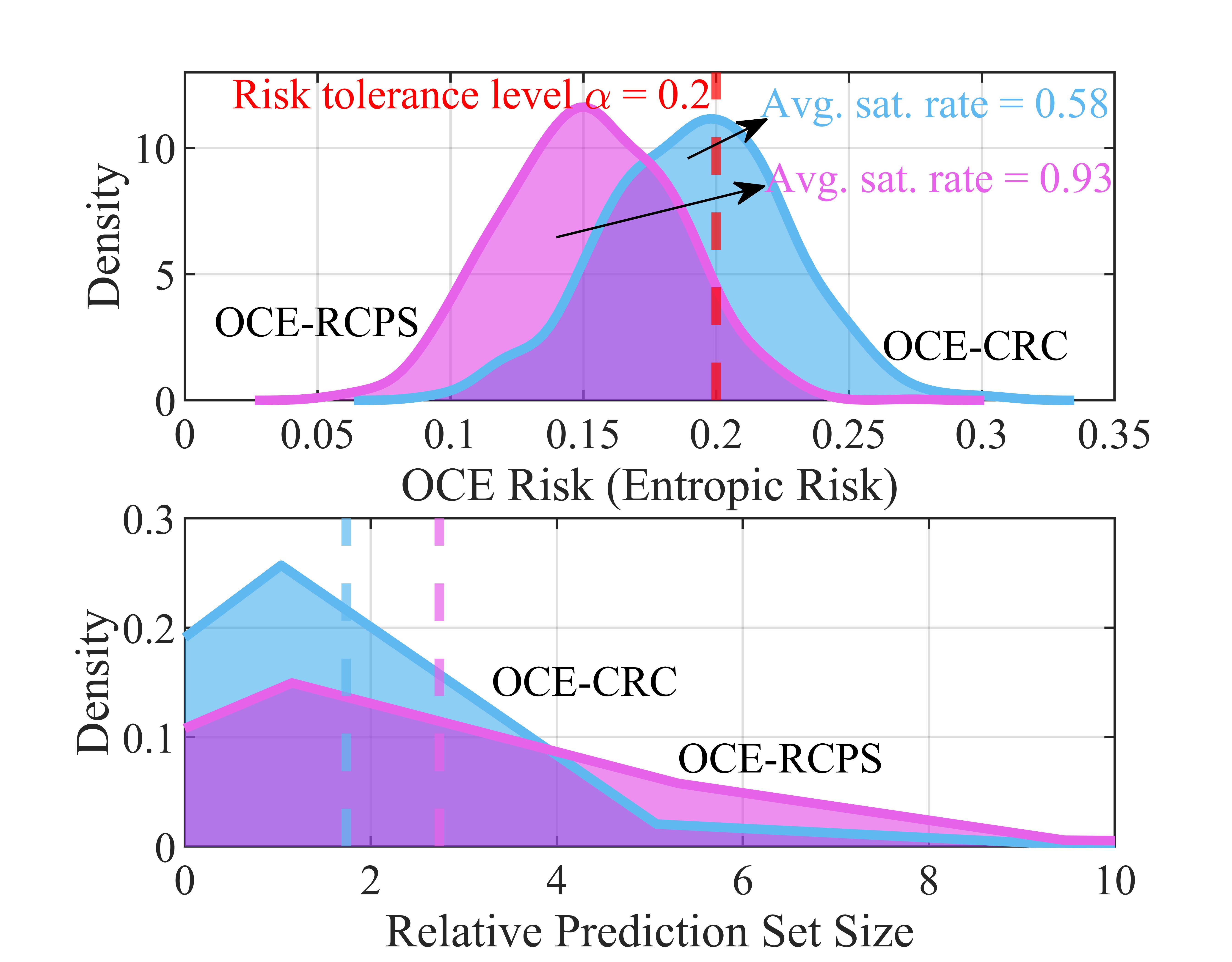}}\vspace{-0.1cm}
     \caption{Distribution of entropic risk (top) and relative prediction set size (bottom) for OCE-CRC and OCE-RCPS, with tolerated risk level $\alpha=0.2$, target satisfaction rate $1-\delta = 0.8$, and entropic risk sensitivity level $\beta=3$. Dashed lines in the bottom panel indicate the median relative prediction set size for each method.}
     \vspace{-0.35cm}
    \label{fig:dist_entropic} 
\end{figure}

\begin{figure} [htb] 
    \centering
    \centerline{\includegraphics[width=\linewidth]{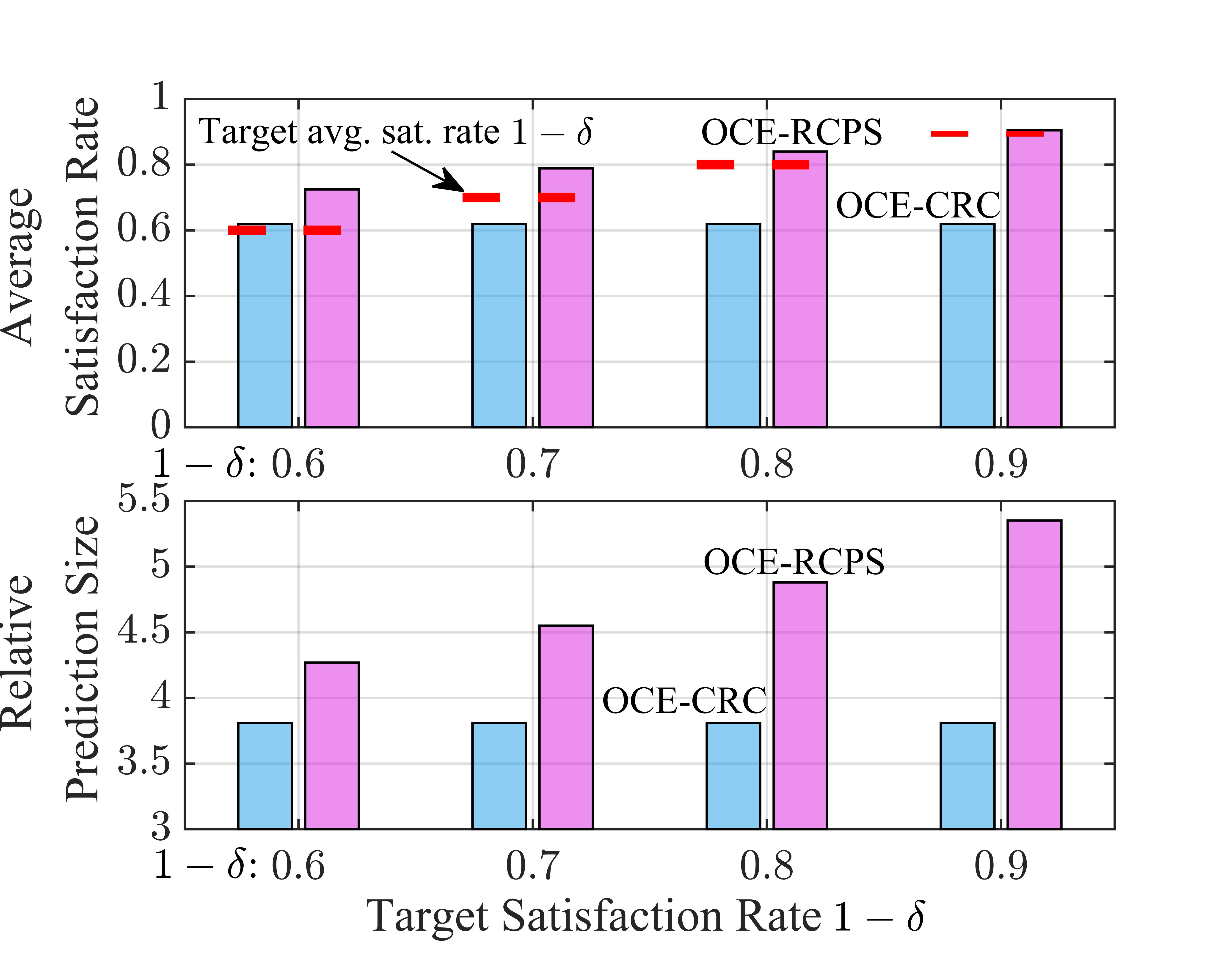}}\vspace{-0.2cm}
     \caption{ Average satisfaction rate (top) and relative prediction set size (bottom) for OCE-CRC and OCE-RCPS across varying target satisfaction rates $1-\delta \in \{ 0.6, 0.7, 0.8,0.9 \}$, with tolerated risk level $\alpha=0.4$ and CVaR concentration level $\beta=0.9$. Red dashed lines in the top panel indicate the corresponding target levels $1-\delta$.}
     \vspace{-0.8cm}
    \label{fig:changing_delta} 
\end{figure}

\noindent \textbf{Robustness of OCE-RCPS across reliability configurations:}
We now turn to analyzing the performance of OCE-RCPS across a range of configurations. The top panel of Fig.~\ref{fig:changing_delta} shows the average satisfaction rate as a function of the target quantile level $1-\delta$. It is observed that, as the target satisfaction rate $1-\delta$ increases from $0.6$ to $0.9$, OCE-RCPS consistently meets the target reliability requirement as in (\ref{eq:oce_rcps_requirement}), while the average satisfaction rate of OCE-CRC remains around $0.6$, irrespective of the choice of the target of $1-\delta$. This confirms that OCE-CRC lacks a mechanism to control the probabilistic requirement on the OCE risk as in (\ref{eq:oce_rcps_requirement}). By contrast, as shown in the bottom panel of Fig.~\ref{fig:changing_delta}, OCE-RCPS provides a controllable reliability knob, adaptively inflating the prediction set size to meet the prescribed target.


\section{Conclusion}
In this paper, we proposed OCE-RCPS, a framework that can control the tail behavior on the OCE risk with respect to the calibration data, not just merely on average. Future directions include extending OCE-RCPS to distribution-shift settings and applying the framework to other safety-critical domains such as autonomous driving.


\bibliographystyle{IEEEtran}
\bibliography{ref}

@article{yeh2025conformal,
  title={Conformal Risk Training: End-to-End Optimization of Conformal Risk Control},
  author={Yeh, Christopher and Christianson, Nicolas and Wierman, Adam and Yue, Yisong},
  journal={arXiv preprint arXiv:2510.08748},
  year={2025}
}

@article{shafer2008tutorial,
  title={A tutorial on conformal prediction.},
  author={Shafer, Glenn and Vovk, Vladimir},
  journal={Journal of Machine Learning Research},
  volume={9},
  number={3},
  year={2008}
}

@article{bates2021distribution,
  title={Distribution-free, risk-controlling prediction sets},
  author={Bates, Stephen and Angelopoulos, Anastasios and Lei, Lihua and Malik, Jitendra and Jordan, Michael},
  journal={Journal of the ACM (JACM)},
  volume={68},
  number={6},
  pages={1--34},
  year={2021},
  publisher={ACM New York, NY}
}

@inproceedings{pogorelov2017kvasir,
  title={Kvasir: A multi-class image dataset for computer aided gastrointestinal disease detection},
  author={Pogorelov, Konstantin and Randel, Kristin Ranheim and Griwodz, Carsten and Eskeland, Sigrun Losada and de Lange, Thomas and Johansen, Dag and Spampinato, Concetto and Dang-Nguyen, Duc-Tien and Lux, Mathias and Schmidt, Peter Thelin and others},
  booktitle={Proceedings of the 8th ACM on Multimedia Systems Conference},
  pages={164--169},
  year={2017}
}

@article{borgli2020hyperkvasir,
  title={HyperKvasir, a comprehensive multi-class image and video dataset for gastrointestinal endoscopy},
  author={Borgli, Hanna and Thambawita, Vajira and Smedsrud, Pia H and Hicks, Steven and Jha, Debesh and Eskeland, Sigrun L and Randel, Kristin Ranheim and Pogorelov, Konstantin and Lux, Mathias and Nguyen, Duc Tien Dang and others},
  journal={Scientific data},
  volume={7},
  number={1},
  pages={283},
  year={2020},
  publisher={Nature Publishing Group UK London}
}

@article{bernal2012towards,
  title={Towards automatic polyp detection with a polyp appearance model},
  author={Bernal, Jorge and S{\'a}nchez, Javier and Vilarino, Fernando},
  journal={Pattern Recognition},
  volume={45},
  number={9},
  pages={3166--3182},
  year={2012},
  publisher={Elsevier}
}

@article{silva2014toward,
  title={Toward embedded detection of polyps in wce images for early diagnosis of colorectal cancer},
  author={Silva, Juan and Histace, Aymeric and Romain, Olivier and Dray, Xavier and Granado, Bertrand},
  journal={International journal of computer assisted radiology and surgery},
  volume={9},
  number={2},
  pages={283--293},
  year={2014},
  publisher={Springer}
}

@inproceedings{fan2020pranet,
  title={Pra{N}et: Parallel reverse attention network for polyp segmentation},
  author={Fan, Deng-Ping and Ji, Ge-Peng and Zhou, Tao and Chen, Geng and Fu, Huazhu and Shen, Jianbing and Shao, Ling},
  booktitle={International conference on medical image computing and computer-assisted intervention},
  pages={263--273},
  year={2020},
  organization={Springer}
}

@inproceedings{guo2017calibration,
  title={On calibration of modern neural networks},
  author={Guo, Chuan and Pleiss, Geoff and Sun, Yu and Weinberger, Kilian Q},
  booktitle={International conference on machine learning},
  pages={1321--1330},
  year={2017},
  organization={PMLR}
}

@article{huang2025calibrating,
  title={Calibrating {B}ayesian learning via regularization, confidence minimization, and selective inference},
  author={Huang, Jiayi and Park, Sangwoo and Simeone, Osvaldo},
  journal={IEEE Transactions on Signal Processing},
  volume={73},
  pages={4492--4505},
  year={2025},
  publisher={IEEE}
}

@article{angelopoulos2022conformal,
  title={Conformal risk control},
  author={Angelopoulos, Anastasios N and Bates, Stephen and Fisch, Adam and Lei, Lihua and Schuster, Tal},
  journal={arXiv preprint arXiv:2208.02814},
  year={2022}
}

@article{ben2007old,
  title={An old-new concept of convex risk measures: The optimized certainty equivalent},
  author={Ben-Tal, Aharon and Teboulle, Marc},
  journal={Mathematical Finance},
  volume={17},
  number={3},
  pages={449--476},
  year={2007},
  publisher={Wiley Online Library}
}

@article{abdar2021uncertainty,
  title={A review of uncertainty quantification in deep learning: Techniques, applications and challenges},
  author={Abdar, Moloud and others },
  journal={Information fusion},
  volume={76},
  pages={243--297},
  year={2021},
  publisher={Elsevier}
}

@article{zhu2025conformal,
  title={Conformal distributed remote inference in sensor networks under reliability and communication constraints},
  author={Zhu, Meiyi and Zecchin, Matteo and Park, Sangwoo and Guo, Caili and Feng, Chunyan and Popovski, Petar and Simeone, Osvaldo},
  journal={IEEE Transactions on Signal Processing},
  year={2025},
  publisher={IEEE}
}
\end{document}